# A Semantic Approach to Summarization

Divyanshu Bhartiya (10250) [1], Ashudeep Singh (10162) [2]

*Department of Computer Science and Engineering, IIT Kanpur*

[1] `divbhar@iitk.ac.in`
[2] `ashudeep@iitk.ac.in`

*Under the guidance of*
*Prof. Harish Karnick*
`hk@cse.iitk.ac.in`

*Abstract*—Sentence Extraction based Summarization of documents has some limitations as it doesn't go into the semantics of the document. Also it lacks the capability of sentence generation which is intuitive for humans. In this report, we hereby take the task of summarization to semantic levels with the use of WordNet and sentence compression for enabling sentence generation. We involve semantic role labelling to get semantic representation of text and use of segmentation to form clusters of related pieces of text. Picking out the centroids and sentence generation completes the task. We evaluate the results of our system against human composed summaries and also present an evaluation done by humans to measure the quality attributes of our summaries.

## I. INTRODUCTION

In today's era, when the size of information and data is increasing exponentially, there is an upcoming need to create a concise version of the information available. Most of the information is available in text form. Till now, humans have tried to create "summaries" of the documents. Creating summaries is a tedious task and one has to read the whole information document to be able to write an abstract and a concise representation of the text. One has to avoid unnecessary and redundant content while creating a summary of a document. One has to get a hold of all the information the document contains and he must be well versed to be able to convey the idea the document represents. There is a need to be able to do the aforementioned task automatically. We hereby propose an approach to do this task of summarization automatically. We came up with a semantic approach to do this task. Our main motivation was to follow the same footprints that a human takes while creating a summary. A human understand the document, and links up the parts of documents that trying to convey a piece of information. He then compresses the information to his need and create the best representation of the document. We hereby introduce the same approach, i.e. identifying the meaning of the document, linking them up, getting the best representation and creating a concise version of it.

## II. PREVIOUS WORK

The earliest work on automatic summarization was around five decades ago. In 1958, Luhn et al. [1] proposed that word frequency plays a very important role in determining its saliency while extracting summaries from scientific text. Subsequently, Baxendale (1958) [2] proposed the importance of a word's position and Edmundson (1969) [3] explained how key phrases can be used to extract summaries. Various works published at those times dealt with the news-wire data or scientific documents. But with the resurgence in the availability of wide tagged corpora and tools available to do processing of natural language at both syntactic and semantic levels, and the ever-growing size of the information available through internet has regrown the interest of NLP researchers in automatic summarization techniques during the last decade.

A crucial issue that the researchers felt the need to address was evaluation. During the past decade, many system evaluation competitions like Text Retrieval Conference (TREC), Document Understanding Conference (DUC) and Message Understanding Conferences (MUC) have created sets of text material and have discussed performance evaluation issues [7]. However, a universal strategy to evaluate summarization is still absent.

Summarization tasks can be widely separated into three classes- Single Document summarization, Multi-Document summarization, Query-based summarization (or topic driven summarization). In single document summarization, the challenge is to look for flow of information or discourse in the text, which should be conveyed by a good summary. While in multi-document summarization tasks, the techniques which are of importance are similarity measures and clustering techniques.

Machine learning techniques for summarization of a single document. Kupiec et al. (1995) [4] described a naive bayes method to determine whether a sentence belongs to a summary or not. For this, they used a large set of technical documents with their abstracts to determine the underlying probabilities of features like sentence length, the presence of uppercase words, presence of more frequent words etc. and then choosing the n highest probable sentences for the summary generation. To this approach, many features were proposed and tested like tf-idf scores etc. But these techniques were soon replaced by other sophisticated machine learning techniques which used bigger feature sets extracted from document and summary.

In later work, Lin (1999) [5] used decision tree arguing that the features extracted from the text were independent of each other. His system performed better than the Naive Bayes systems, this justified the dependence between the various textual features used in the techniques then. During the same time, WordNet [6] became the prime source of deep natural language analysis. For summarization, McKeown and Radev, (1995) emphasized on using semantic structure of text rather than statistics of words of the documents. They describe a notion of cohesion in text, meaning greater cohesion amongst parts of text where similar meaning words occur. The cohesion phenomenon occurs at the word sequence levels and hence form lexical chains. Identifying lexical chains in the text, they used the strongest few lexical chains to represent the text in the summary.

Multi-document summarization requires higher abstraction and information fusion. The earliest attempt on multi-document summarization was SUMMONS (Radev and McKeown 1998). It tackles news articles relating to single domain but collected from varied sources. It has built in message understanding templates for text understanding and representation. The two way step is first processing the input texts and filling in the templates and then using sophisticated linguistic generator to get a single short summary. But as expected the system despite being promising, was not widely used because of narrow domain application and hard-coded templates. This tool was then improved by Barziley (1999) [8] and McKeown (1999) [9], defining a clustering problem that will deal with the text representation problem. To construct clusters they define features like tf-idf scores, noun phrases, proper nouns, and synsets from the WordNet. After themes as identified into clusters, the next step is information fusion. Depending on the dependency parse structures of the extracted sentences/phrases in the themes, specific phrases are used for the final generation part. Once the summary content is decided, a grammatical text is generated by translating those structures to arguments required by FUF/ SURGE language generation system [10]. Through the SUMMONS system and its further improvements, Radev and McKeowns brought in the concept of centroid based clustering into the summarization paradigm, which was then explored by discovering more features that prove important to make two portions of text similar [11].

Although the research on summarization was started about a 50 years ago, there is still a long way to go before systems are built which create beautiful short summaries as good as human generated ones. The research in this area is hot because of the ever increasing information on the World Wide Web and accurate summarization techniques are certainly important for text understanding. With the highly efficient and accurate systems to tag, represent and analyze texts in English like Penn Tree Bank [12] trained POS taggers, WordNet, Named Entity taggers, Semantic Role Labelers etc., there is a hope in the research community that this research in summarization and related fields in Natural Language can be taken through separate directions. In this project, we work on applying these state-of-the-art techniques over text and extracting meaningful and concise summaries in the domain of single-document summarization tasks.

III. TECHNOLOGIES

Our approach to Summarization begins with the task of Pronominal Resolution, followed by Part of Speech Tagging and Semantic Role Labelling, to create frames of a sentence. Wordnet is then used to get to the synsets and the semantics of the sentence. Clustering is then applied to the frames, followed by getting the respective centroids of the frames.

A. *Pronominal Resolution*

Pronominal or Anaphora resolution is an important task before we go deep into the process of summarization. It is widely used in machine translation, summarization or question-answering system. Anaphora is the act of referral that is, it denotes to the antecedent in the left. Anaphora resolution is the task of identifying the subject (noun) for a reference (pronoun). When humans perform pronominal resolution, they keep in mind the central idea and topic of the current information being conveyed. Humans don't need to go back in the document structure or jot down the points to do the resolution. Humans are capable of doing that intuitively and by refreshing the memory they can attain that task. Pronominal resolution is needed to avoid reference structures.

*John helped Mary. She was happy for the help provided by him.*

Now sentence two is more informative than second one. However if the summary only contains sentence 2, that won't make any sense and will be incomprehensible. However on performing the pronominal resolution, we can obtain:

*John helped Mary. Mary was happy for the help provided by John.*

B. *Part of Speech Tagging*

Part of Speech tagging or POS tagging is the task of grammar tagging of a language. It is the task of identifying the nouns, verbs, adjectives, pronouns, adverbs, etc. This is accomplished by tagging each word or token of the text with its respective part of Speech. The tagging is mostly performed in accordance with the grammar rules of language. We use the Stanford's implementation [13][14] of POS tagging to achieve our task. This software uses log-linear part-of-speech taggers. Following is an example to illustrate its use:

*Mary was happy for the help provided by John.*

The output of Stanford POS tagger is a parse tree, which starts at S (sentence) being the top level node and branching arising from the grammar rules of English language. The output is:

```
(ROOT
  (S
    (NP (NNP Mary))
    (VP (VBD was)
      (ADJP (JJ happy)
        (PP (IN for)
          (NP
            (NP (DT the) (NN help))
            (VP (VBN provided)
              (PP (IN by)
                (NP (NNP John))))))))
    (. .)))
```

**Fig1. POS Tagger Example**

This parse tree will help us retrieve the nouns and the verbs in a sentence which will be put to further use.

*C. Semantic Role Labelling*

Semantic role labelling, is a shallow semantic parsing technique in NLP. It detects the predicates associated with a verb in a sentence. It is the task of finding the arguments and the modifiers associated with a verb. It is analogical to a function with certain parameters. Each function can be considered a verb corresponding to an action. As each action is associated with an agent and a theme, the parameters of the function can be considered as the agent and themes. Each verb is associated with modifiers like temporal, locational or an adverb. These modifiers can also be considered to be parameters of the respective function representing the verb. So in short:

*If a sentence is represented by the following pattern,*
*<Agent> <action> <theme> <modifiers>, then the sentence can be translated as F (arg1, arg2…argN) where F is the <action> and <arg1>…<argN> are the <agent>, <theme> and <modifiers> respectively.*

Example:
     *[ARG0 John] helped [ARG1 Mary];*

Semantic Role Labelling captures the semantics of a sentence, as this help retrieve the specific actions and their respective arguments. Semantic role labelling helps in finding the meaning of the sentences and the associated actions in the sentence. It recognises the participants of the propositions governed by verbs,

- **SENNA**: It is a software which does a variety of NLP tasks, like POS tagging, NER recognition, Chunking, Semantic Role Labelling and Syntactic parsing [15]. It's a C based system which we employed to do semantic role labelling. Given a sentence, SENNA SRL creates frames for each of the verb occurring in the sentence.
- **PropBank Annotation**: "PropBank is a corpus in which the arguments of each predicate are annotated with their semantic roles in relation to the predicate [16]." PropBank associates each verb with arguments and some modifiers. It is similar to FrameNet, each of this predicate is associated with a frame.

E.g. *"Mr.Bush met him privately, in the White House, on Thursday."*

*Relation: met*
*Arg0: Mr.Bush*
*Arg1: him*
*ArgM-MNR: privately*
*ArgM-LOC: in the White House*
*ArgM-TMP: on Thursday.*

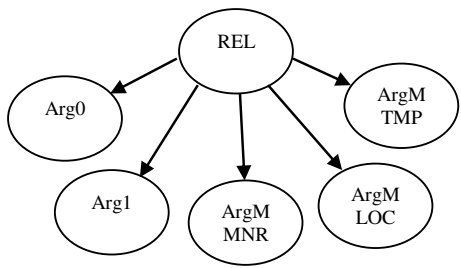

**Fig 2. Structure of a frame by SRL**

We use these frames to our disposal to get the semantic representation of our document. Each frame represents a part of sentence describing its meaning.

*D. WordNet*

WordNet is a freely available standard corpus from Princeton University [6], put to the use of Natural Language tasks. WordNet is a lexical database of English, comprising and arranged in Nouns, Verbs, Adjectives and Adverbs. They are grouped into sets of cognitive synonyms. These sets are called Synsets as they describe the semantic and lexical relations between words. Similar types of words form a synset, meant to convey a similar sense of meaning. Synsets are organized in relations based on their hierarchy. Noun synsets have a relationship of hyponymy and hypernymy. For example: car is a kind of vehicle, then the synset of car has a hypernym synset that contains vehicle. Similar holds for hyponymy. Verbs synsets have relationship of hypernymy and troponymy. Troponymy forms the bottom of the tree and expresses more specific manners. We use WordNet to get the synsets of the nouns and verbs associated with a frame. Wordnet will enable to find the related pieces of text in a document and will ease up the problem of textual entailment. The different levels of hierarchy lead to different levels of abstraction. In our approach we just to one level up and one level down search, not going deep into relationships of hyponymy or hypernymy.

## IV. OUR APPROACH

Summarization task was initiated with the thought in mind of getting a summary of the document which will not be based on extraction of informative sentences from the document, but the idea of generation of sentences. To create summaries, we should be able to represent the document in a model from which it's possible to generate sentences. Also, we needed the semantics to come into play, while creating the summaries. So our idea of generation of sentences comes from compressing a given sentence. This can be achieved in a lot of fashion, for example getting rid of unnecessary verbs from a sentence or avoiding adverbs.

Our system takes a document as an input and does the pronominal resolution on it. Pronominal resolution is used to form chains in the documents and resolve the pronouns with their respective subjects. This will help us in the process of extracting information chunks without any ambiguity. Replacing pronouns with their respective and related nouns will help us in the later phases were the measurement of context or flow of information is difficult to manage. Getting rid of ambiguous or questionable information is a prior task in our approach of summarization.

Once the document is resolved we do the part of speech tagging on it. Part of speech tagging is a basic move in natural language processing. Identifying nouns and verbs is the main motivation of doing POS tagging. POS tagging forms the basis of semantic role labelling, where frames are built on verbs as the root of the frames. Also the use of part of speech tagging enables us to identify the elements on which the synsets need to be constructed.

Now we move to the crux of our system. Now we perform semantic role labelling on the pre-processed data. Semantic role labelling enables us to reduce the granularity of the document. The main motivation of using SRL is to introduce semantics. An approach was needed to decrease the granularity of document from sentences to a model, where the semantics of the sentence can be represented. In short, we needed a representative model which will clearly depict the semantics. SRL functionality of SENNA helps us to create frames from a sentence [15]. Each frame will be constructed using verb as the root node and its respective arguments as its branches, hence a PropBank annotation. Hence, a sentence containing one or more verbs is now a collection of frames.

If document D has sentences $S_1, S_2 \ldots S_N$, after this phase, we have a collection of frames $F_1, F_2 \ldots F_M$, where $M>N$. Now we don't need sentences to work around with, and here on we progress on using these frames as the basis.

$$D = \bigcup SRL(S_i) \; \forall \; i \in N$$

Now that the document is decomposed into frames, we move forward to the next phase of forming synsets. WordNet dictionary helps to find the synsets, the hyponym synsets and hypernym synsets. This is used for the fact that a meaning can be later be repeated in the document, and hence to avoid redundant data, WordNet can be used to capture this. WordNet will help to collect all the pieces of text in the document where a similar idea or type of information is described. This will enable us to capture all those sentences where the topic of description remains same. To go about this idea, we find the hyponym and hypernym synsets for the arguments' nouns. For the frame's root verb, we find the hypernyms and troponyms of the verb. Now each frame is associated with two sets, a set describing the arguments' synsets and a set describing the verb's synsets.

$$Frame_i.nounSynsets = \bigcup Synsets(n), \forall n \in Frame_i.Args$$
$$Frame_i.verbSynsets = Synsets(v), v = Frame_i.root$$

Where,
$$Synsets(n) = Hyponyms(n) \cup Hypernyms(n)$$
$$Synsets(v) = Hypernyms(v) \cup Troponyms(v)$$

Each frame now describes the possible meanings and also have a sematic representation to it. Now we do the task of textual entailment where we use the synsets created by WordNet to find frames entailing each other. This is accomplished by finding the match between a frame's arguments' synsets and another frame's arguments' synsets. This measures the entailment content of an agent or theme of a frame with another. In short, if there is high matching between the two frames' arguments' synsets, it reveals that the subject of two frames or the parameters of two different frames' verbs are same. We also measure the matching between the verb's synsets to see if the similar type of action is performed somewhere else in the document. Let the score of arguments' synsets matching be denoted by $A(F_i, F_j)$, that is it calculates the arguments' synsets similarity. Similarly that matching score of verb's synsets is depicted by $V(F_i, F_j)$. The minimum scores is 0, which is nothing matches at all. These two matchings lead to various scenarios in order of priority where the frames' meaning are measured.

- $A(F_i, F_j) \neq 0$ and $V(F_i, F_j) \neq 0$: this is the highest priority case. The scores reveal that there is some resemblance between two frames in terms of the arguments as well as the action. Both the frames are talking about similar agents doing some similar action.
- $A(F_i, F_j) \neq 0$ and $V(F_i, F_j) = 0$: this is the second scenario where there is no matching between verb's synsets. This case tells that the two frames have similar parameters but there is no relationship between the actions performed by the arguments. Both the frames talk about similar agents but completely different tasks are performed by them, for example "$F_i$: John helped Mary. $F_j$: John killed Bob."
- $A(F_i, F_j) = 0$ and $V(F_i, F_j) \neq 0$: this is the third priority case, where the two frames talk about similar actions but their agents are different. This scenario is of lower priority because subject matching is more important than action matching, for example "$F_i$: John helped Mary. $F_j$: Bob helped a dog".

- A $(F_i, F_j) = 0$ and V $(F_i, F_j) = 0$: this is the lowest priority matching. The frames are irrelevant to each other and have no sort of connection between them

Now once we have the matching scores, we can link these frames as based on their matching score. Construct a graph G with frames being the nodes of a graph and the edges being the matching between the two frames. The graph will be a directed weighted graph, the wrights being the sum of A $(F_i, F_j)$ and V $(F_i, F_j)$. This score measures the similarity between two frames. Constructing the graph will connect all the related frames. We take only a limited number of matchings based on the score of matching. We reduce the number of edges, taking some of the top scores to create edges. This ensures we connect only those frames that have good similarity measure score.

**SEGMENTATION**: We introduce the process of creating segments in the graph. This is accomplished by joining all the connected frames from a source node. It basically is creating of all frames that are accessible from an originating node. The algorithm for creating segments is:

*S = {}//set of all segments*
*k=0          //number of segments*
*For each frame f in document:*
    *If (f ∈ $S_j$) for some j*
        *Add f.edges to $S_j$*
    *Else*
        *Create a new segment $S_k$*
        *S=S ∪ $S_k$*
        *k++*
        *Add f to $S_k$*
        *Add f.edges to $S_k$*
*End*
*Return S*

**Algortithm1. Creating Segments**

Segmentation generates clusters in our graph i.e. it forms all the frames that have some sort of relationship among each other. We can say that the clusters so obtained contain similar type of information and those have frames have a great deal of matching amongst each other. Now the problem just remains to get the frames that will hence be the best representation of the frames of the segments so thus formed. We can say these frames to be a good and concise representation of the information depicted by the frames in the cluster.

**Centroids Extraction:** Just like the centroid of a cluster are the middle point of a cluster, we call the best frames of a segment to be the centroid of that segment. Getting the centroids from a segment is based on a number of features like the number of incident on that frame, the number of outgoing edges from that frame, the frame position in the document, the length of the frame, the number of named entities in the frame, etc. We calculate a weighted sum of these features for each frame in a cluster.

$$frame.featureScore = \sum w_i * f_i, \forall f_i \in frame.features$$

The score obtained from incoming and outgoing edges is of great value as it tells us about the information network present in the document due to that particular frame. Each frame gets a score based on these features and the frame with highest scores are chosen as the representation of the segment. We extract some number of frames from each segment, and hereby we obtain the final frames that will represent our document. To generate sentences from these obtained frames, we observe the arguments in the frame and generate sentences by concatenating [Arg0] [verb] [Arg1] [Arg2]. The sentences so obtained are expected to be a summary of the document input. We take the assumption that [Arg0] is the subject of [verb] and [Arg2] represents the object of [verb]. This assumption does not always lead to grammatically correct sentences.

V. **OBSERVATIONS AND RESULTS**

Semantic Role labelling receives all transitive verbs and creates frames out of it. Since the transitive verbs represent some action that is performed and have a subject and an object, intransitive verbs lack these qualities. They lack the need of an object to which the verb is referenced. Hence SRL does correct labelling for most of the sentences in the document. It is also able to recognise most of the verbs that are associated with a proposition. Finding similarity on the basis of WordNet gives satisfying results. Apart from synset matching, we employed the use of individual matching of words present in verbs and arguments, to avoid repeated sentences that hold the same meaning, and thus avoiding redundant information. Feature score on the basis of number of outgoing and incoming edges result the best score as it links up all the connected frames, henceforth making it the most suitable frame to be picked for representing the frames it has joined. The features taken for calculating the centroids helps in getting impressive frames from all over the document. The process of segmentation and getting centroids produces good quality frames.

However the sentence generation part is a little faulty as it has not been able to produce grammatically correct sentences. Also sometimes the centroids that are selected, are from the clauses of the sentences they represent. This means, that these frames may not be able to generate grammatically correct sentences, since they don't from the principal verb of the sentence. Some of the sentences are grammatically correct and don't require modification. The sentence generation needs improvements to be able to recognise the type of sentence it can form from the frame so obtained.

To measure the performance of our summary we use a document which already has a human written summary S. We measure the performance of our summary generated, say S` with S. To measure the summary quality, we perform the frame formation process S`. This will result in a collection of frames. Now we have two collections of frames, that is the frames (centroids) of our segmentation process on the document, and the set of frames obtained from doing semantic role labelling. We again perform similarity matching using WordNet between these two sets.

$$\text{Sim}(S, S') = \sum_{i \in S.\text{Centroids}} (\text{argmax}_{j \in S'.\text{frames}} \text{sim}(i, j))$$

The similarity score between the two summaries depends upon the number of centroids we choose to form our summary. The average score comes out to be about 50% match. This is not a bad result as we can't expect the summaries to be of the same intellectual and abstract quality. Our approach focussed on semantics and sentence compression. This result shows that the semantics are well conveyed in our summary. To measure the qualities of a summary like expressiveness, information content, abstraction we performed a human evaluation.

Human Evaluation: We executed our system on some chapters from "Into the Wild", "Animal Form" and "NCERT Class X Science". We had summaries on those chapters beforehand. We asked humans to rate our system generated summaries in reference to human compiled respective summaries, on the grounds of Information Content, Grammatical Correctness, Abstraction, Expressiveness and Excess or Unnecessary Information. These qualities are rated on a scale of 1 to 5, 1 being the least and 5 being the most. The mean and standard deviation observed over these qualities is depicted in the table. The results suggest that information content is good in our summaries generated. But some people are of the opinion that there is a lot of unrelated information, and the standard deviation varies a lot. Also we need to improve upon the abstractness quality, as most of the frames are just a short hand representation of sentences in document and hence turn out to be factual.

| S.No | Summary Quality Attributes | | |
|---|---|---|---|
| | Quality | Mean | Standard Deviation |
| 1 | Information Content | 3.81 | 0.96 |
| 2 | Grammatical Correctness | 3.69 | 0.87 |
| 3 | Abstractness | 3.35 | 1.00 |
| 4 | Expressiveness | 3.62 | 0.84 |
| 5 | Excess/Unnecessary Detail | 3.19 | 1.30 |

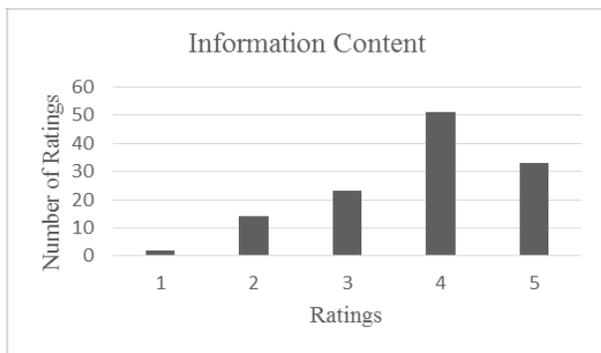

**Fig. 3**

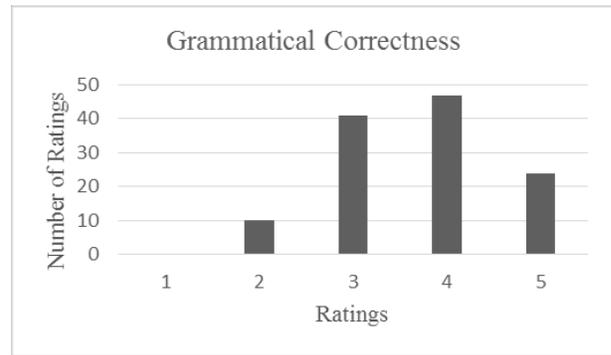

**Fig. 4**

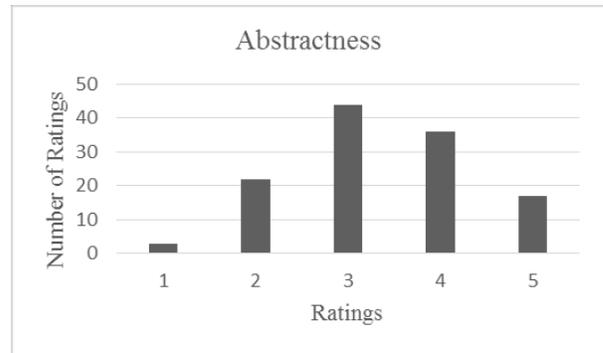

**Fig. 5**

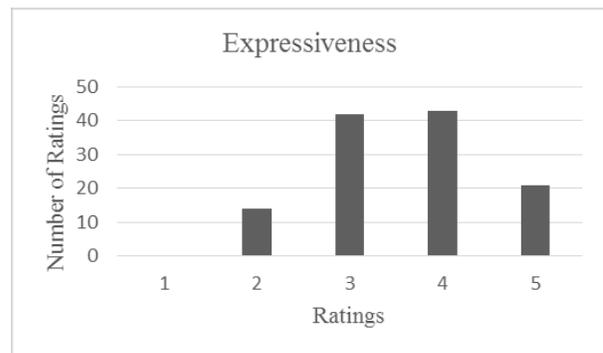

**Fig. 6**

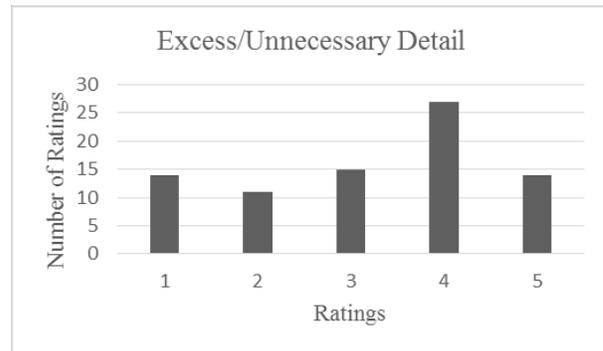

**Fig. 7**

## VI. CONCLUSION

Summarization is an important field of research these days and finds applications in the fields of information retrieval like in news mining [17] and customer review mining [18] and educational systems like intelligent tutors [19], question generation systems [20]. Many different approaches have seem to work well on different types of text, like factual, scientific papers, or news. Our system seems to perform well on factual data. Further modifications could be improving the feature space for centroid selection, better sentence generation capabilities, or improved metric for semantic similarities. Also a lot number of heuristics can be applied to preprocess the document and improve the quality of information in the primary stages. Our system needs improvement in sentence generation and semantic role labelling. This work has been done in the motivation to produce better quality and meaningful summaries, which has been exemplified by our results.